\pdfoutput=1
\documentclass[11pt]{article}

\usepackage[final]{acl}

\usepackage{times}
\usepackage{latexsym}

\usepackage[T1]{fontenc}
\usepackage[utf8]{inputenc}

\usepackage{microtype}

\usepackage{custom}

\title{Zero-shot prompt-based classification: topic labeling in times of foundation models in German Tweets}

\author{Simon Münker \\
  krAil, University Trier \\
  \texttt{muenker@uni-trier.de} \\\And
  Kai Kugler \\
  krAil, University Trier \\
  \texttt{kuglerk@uni-trier.de} \\\And
  Achim Rettinger \\
  krAil, University Trier \\
  \texttt{rettinger@uni-trier.de} \\
}

\begin{document}
\lstset{language=Python}

% Title Page
\maketitle

% Body
%
% -----------------------------
\begin{abstract}
Filtering and annotating textual data are routine tasks in many areas, like social media or news analytics. Automating these tasks allows to scale the analyses wrt. speed and breadth of content covered and decreases the manual effort required. Due to technical advancements in Natural Language Processing, specifically the success of large foundation models, a new tool for automating such annotation processes by using a text-to-text interface given written guidelines without providing training samples has become available. 

In this work, we assess these advancements \textit{in-the-wild} by empirically testing them in an annotation task on German Twitter data about social and political European crises. We compare the prompt-based results with our human annotation and preceding classification approaches, including Naive Bayes and a BERT-based fine-tuning/domain adaptation pipeline. Our results show that the prompt-based approach – despite being limited by local computation resources during the model selection – is comparable with the fine-tuned BERT but without any annotated training data. Our findings emphasize the ongoing paradigm shift in the NLP landscape, i.e., the unification of downstream tasks and elimination of the need for pre-labeled training data. 
\end{abstract}
%
% -----------------------------
\section{Introduction}

Since the release of ChatGPT in November 2022, the focus of both public and scientific interest has shifted to generative NLP technologies like Large Language Models (LLMs). Key are questions on human-machine interaction, specifically what benefits such tools can bring for automating manual tasks. Generative foundation models can act like multilingual chatbots \citep{ouy22}, which follow instructions given as natural language text while interpreting texts by statistically capturing human knowledge and replicating human language understanding capabilities. Structuring and formulating these commands has been termed \textquote{prompt engineering} that, in conjunction with powerful models, makes it possible to solve tasks the model has not been (extensively) trained on. Such a capability of models is known as zero- or few-shot learning.

When combined, those three capabilities, i.e., instruction-following, natural language understanding, and few-shot learning, promise a new level when automating annotation processes of textual data concerning the manual effort required. Like human annotators, those systems only need to be provided with annotation guidelines as part of their prompts to replicate human annotation behavior. The difference is that LLMs never tire when processing textual content and scale in numbers.

The approach fits well with the needs of researchers interested in investigating current topics of interest as expressed in online social networks. As the frequency of societal crises increases \citep{gut22}, a timely analysis is crucial to understanding tipping points in public opinion. Projects that rely on surveys like SOSEC\footnote{SOSEC Project Homepage: \url{https://www.socialsentiment.org}} consult their survey participants weekly to keep up with the news. However, even weekly updates might not be fine-grained enough to capture influential events. Here, LLMs potentially offer a complementary tool that can keep up with the temporal and quantitative scale needed for high-frequency analytics.

In this work, we investigate the possibilities of using an open pre-trained generative language model to process a social media text dataset as it would be observed \textit{in-the-wild.} The requested annotations are shown to be challenging even for human annotators, despite providing extensive annotation instructions. Here, our main focus is not on building a better annotation approach with respect to overall accuracy, but on how well current LLMs can serve as an automated primary annotation tool without being shown examples. We specifically assume an experimental setup that requires open local models for control and reproducibility reasons with moderate hardware requirements.

Accordingly, we address the following research questions:

\begin{description}
    \item[RQ$_1$] Can zero-shot prompt-based classification achieve comparable results to a fine-tuned classifier and align well to human annotations? 
    \item[RQ$_2$] How does the scope of information provided to the model, i.e.~the extent of annotation guideline impact the performance? 
\end{description}

In in addition to answering our research questions. We provide a standalone Python module for prompt-based classification with local LLMs (see Sec. \ref{sec:methods:implementation}).
%
% -----------------------------
\section{Background}

The motivation for our work is twofold. Content-wise, the political and social situation in the EU poses a relevant interdisciplinary subject. In particular, how citizens express their opinions on online social media platforms. For the scope of this work, we omit a detailed description. Collecting large amounts of unlabeled data comes with the need for annotation to enable future analysis. Streamlining the annotation displays our technical motivation. With the advent of LLMs capable of performing various tasks, new approaches emerged to classify textual data. Notably, methods allow classifying content through a text-2-text interface, where the user can align the classification expectations based on textual annotation guidelines. That omits the need for machine-learning-based optimization and shifts the focus to formulate human-readable guidelines that the model can follow.

Text classification, like sentiment analysis or topic labeling, holds significant importance in both research and the economy \citep{pet23}. It enables us to extract valuable insights from textual data and make informed decisions across various domains, including customer feedback analysis, market research, and automated content moderation \citep{min21}. Traditionally, text classification relied on supervised learning approaches utilizing task specific models \citep{kad19} or fine-tuning a pre-trained models on a labeled datasets \citep{weis23}. The development of optimized and robust text classifiers is therefore a resource-intensive task.

\paragraph{Instruction Fine-tuning}
The success of LLMs was followed by a paradigm shift triggered by a proposal from Google in 2020 \citep{raf20}, \citep{sun22}. To this point, the typical pipeline combines fine-tuned models like BERT \citep{vas17} or XLNet \citep{XLNET} with a task-specific classification head. For classification tasks, the attached head architecture produced a probability distribution over the given classes \citep{kan18}. For generative tasks, a sequential decoder was used as an attached head, which generates a text sequence as output \citep{jia21}. In contrast, the unified pipeline has three main advantages: a) the optimization pipeline, including the data preparation, is more efficient as the models achieve state-of-the-art performance with less labeled data, b) the approach strengthens the capability of transferring knowledge to unseen tasks using a known formulations, and c) from the non ML researchers perspective, unified models are easier to infer and deploy.

\paragraph{Prompt Engineering}
Instruction-based model solve tasks that are provided in human-like text during conversations. However, the effectiveness of these models relies heavily on the quality and specificity of prompts given to them. Prompt engineering, the process of formulating and refining prompts, plays a crucial role in harnessing the full potential of LLMs \citep{liu21}. Unlike the traditional pipeline for supervised tasks, which trains a model to take in a textual input and predict an output, prompt-based approaches utilize LLMs in a dialog.

This paradigm shift allows us to bypass the aforementioned bottlenecks. We no longer require pre-labeled datasets for fine-tuning the models specifically for each application. Instead, we can utilize the model's general language understanding capabilities and prompt it with task-specific instructions. This significantly reduces the need for large-scale labeled datasets \citep{sun22}, which can be expensive and time-consuming to create.
%
% -----------------------------
\section{Data}

To assess the capabilities of zero-shot prompt-based classification \textit{in-the-wild} we deliberately did not resort to an academic benchmark, since they tend to not reflect the challenges of real-world topic labeling appropriately. Also, we intended to avoid a standard but unrealistic setting with English only data.

% -----------------------------
%
\subsection{Collecting}

We collected a German Twitter data set according to a topical selection defined by the survey questions of the SOSEC project about the energy crises in the winter of 2022/2023. The non-English data set was picked to further stress-test the LLMs' capabilities in a realistic setup. At that time, Twitter (now $\mathbb{X}$) still provided API access. We compiled a comprehensive list of hashtags and keywords that broadly reflected the described crises. The list consisted of relevant terms, including trending keywords, hashtag-based identifiers of political parties, and persons of interest. We queried for each keyword in the list consistently between October 2022 and May 2023. During this time, we collected approximately 750,000 samples.

% -----------------------------
%
\subsection{Manual Annotation}

Two domain experts and native speaker annotated a random selection of approx. 7000 tweets. The annotators were instructed accordingly and given a manual with guiding questions on whether a tweet should be annotated or not. Of the selection samples, only 3000 could be annotated as belonging to a topic, as many tweets did not match our criteria. A high degree of noise due to ambiguity, variation, and uncertainty is a common property of real-world data sets \citep{bec20}.
%
% -----------------------------
\section{Methods}

The candidate methods we picked for automating the annotation task, are taken from three eras of modern NLP: A Naive Bayes classifier, representing the pre-deep learning era, is picked as the baseline. Next, for the deep learning era, a pre-training and fine-tuning approach using a BERT transformer \citep{BERT} is selected. Finally, for the era of foundation models, we use instruction-tuned models based on the transformer T5 \citep{T5}. Again, we tried to setup a realistic ''in-the-wild'' scenario by picking freely available models, that can be run on moderate hardware requirements.

% -----------------------------
%
\subsection{Baseline}
In order to establish a baseline for the methods and our prompt-based classification task, we employ a Multinomial Naive Bayes Classifier \citep{man09}. To represent our text data numerically, we utilize a count vectorizer also provided by scikit-learn \citep{scikit}. The count vectorizer converts the textual data into a matrix of token counts, where each row represents a sample, and each column represents a unique word or token in the corpus.

% -----------------------------
%
\subsection{Fine-tuned Transformer}
We chose the  model \textquote{gbert-base}, for German BERT, which is a language model specifically designed for text classification and Named-entity recognition in German \citep{cha20}. For our tasks, we fine-tune all parameters on 80\% of the annotated data as a single class classification task. Upon completion of the model development and training, we deployed the models to the Hugging Face model hub. The models are available under the \textquote{cl-trier}\footnote{Models available on Hugging Face: \url{https://huggingface.co/cl-trier}} account, allowing other users to access and utilize them for their own applications.

% -----------------------------
%
\subsubsection{Additional Domain Adaption}
To further improve the performance of our fine-tuned classification model, we utilize our raw data (approx. 750,000 tweets). Thus, we include an additional pre-training phase to shift the model's language understanding toward the target domain \citep{ram20}. We shift the focus of the generalized pre-trained BERT model to a Twitter-specific language. That improves the robustness of the model to achieve out-of-distribution generalization without training a model from scratch for our task. The inclusion of a second pre-training phase (adaptive pre-training) improves performance and generation significantly for classification tasks \citep{man22}.

% -----------------------------
%
\subsection{Zero-Shot Prompting}
The two preceding methods set the traditional machine-learning baseline and current SOTA for text classification. Our text-to-text zero-shot prompting \citep{koj22} approach differs in two main aspects. It benefits from the text input and text output paradigm and, thus, pulls away from mathematical optimization. Thereby, we can study the impact of textual formulation on our annotation goal, align the annotation by words, and not optimize by parameters. It does not rely on training data or examples (zero-shot) and, thus, cannot overfit the provided data or assimilate the included biases.

We restrict our setup and the model selection to a level that modern desktop workstations (5.000 Euros in 2023) can effectively run the program. With this, we underline the applicability during active research for smaller groups or individuals. For our experiments, we compare a monolingual and a multilingual instruction-tuned model in four different sizes. Regarding the prompts, we analyze the performance of levels of textual detail, from vague introductions to a reduced version of the annotation guidelines.

\paragraph*{Model Selection}
To allow for a reproducible experimental setup we limit our selection to freely available models from the platform Hugging Face supporting English and German and trained in an instruction-tuned text-to-text scenario. With this filter, the selection is reduced – selection date: Mai 2023 – to two models, namely Flan-T5 \citep{FLAN-T5} and mT0 \citep{mt0}. Both models are based on the same fine-tuned transformer T5, each fine-tuned and adapted in a custom manner. This selection allows for a comparison and evaluation of the adaption quality beyond prompt templates alone. Both models are available in four different sizes, usable with our restrictions. Thereby, we can compare, in addition, the respective performance across several parameters. It gives us a third dimension of analysis.

\paragraph*{Prompts} We provide a baseline prompt (Prompt \ref{prompt:base}) that is generic without a specific task description. The terms in curly braces represent variables, substituted during prompting. To differentiate the task description from the text content, we use triple back-ticks ($'''$) as delimiters \citep{whi23}. Additionally, the template emphasizes choosing a single class through the keywords \textquote{categorize} and \textquote{one of}.

\begin{prompt}[ht]
	\begin{lstlisting}[breaklines]
prompt: str = f"""
Categorize the following tweet into one of the listed classes {classes}:
'''{text}'''
"""
    
classes: List[str]
text: str
	\end{lstlisting}
	\caption{base}
	\label{prompt:base}
\end{prompt}

In the preceding prompt, we omit a naming type of classification task. In the following prompts, we gradually add levels of information. To analyze if and how the models benefit from an additional explanation. In the first prompts, we introduce the name of the respective tasks (Prompt \ref{prompt:task-name}). As both models are fine-tuned for various classification tasks, we assume that they benefit from the task names.

\begin{prompt}[ht]
	\begin{lstlisting}[breaklines]
prompt: str = f"""
Your task is to classify the following tweet regarding its (topic|sentiment) into  one of the following classes {classes}:
'''{text}'''
"""
	\end{lstlisting}
	\caption{task-name}
	\label{prompt:task-name}
\end{prompt}

In the following two prompts, we give a short description about the task. In addition to naming the task explicitly, we provide additional synonyms for task (Prompt \ref{prompt:description:topic}.

\begin{prompt}[ht]
	\begin{lstlisting}[breaklines]
prompt: str = f"""
Your task is to analyze the topic of the following tweet:
'''{text}'''
Thus, identify the dominant subject of the tweet content and classify it into one of the following classes: {classes}
"""
	\end{lstlisting}
	\caption{description, topic}
	\label{prompt:description:topic}
\end{prompt}

The last prompt we tested contain a condensed version of the annotation handbook (Prompt \ref{prompt:handbook:topic}. We could not use the full version as our models are restricted in the input length, and the complete topic task description would not leave room for the input of the tweet. With this information, we provide the model with nearly identical instructions as the human annotators.

\begin{prompt}[ht]
	\begin{lstlisting}[breaklines]
prompt: str = f"""
Your task is to utilize the following class descriptions, label between *'s followed by its definition to choose the one most fitting for the tweet:
*Wirtschaft*: The tweet contains concerns about the economic crisis or the personal financial situation.
*Migration*: The tweet evaluates the chances and dangers of migration and makes judgmental remarks about migrants or as migrants perceived people.
*Demokratie*: The tweet expresses trust or distrust towards the parliament and advocates or rejects the democratic system.
*Ukraineunterstuetzung*: The tweet states the author's position on the Russo-Ukrainian war, evaluates economic penalties against Russia, or postulates financial or military support for Ukraine.
*Energiewende*: The tweet discuss personal concerns about the power supply or energy system transformation.
'''{text}'''
""" 
	\end{lstlisting}
	\caption{handbook, topic}
	\label{prompt:handbook:topic}
\end{prompt}

\paragraph*{Metric}
In traditional machine learning classification pipelines the model response represents one of the given classes or numerical representation. However, in our prompt-based approach, the models respond with unrestricted free-form text. Thus, the model is not limited to responding with one of the targets but may produce additional explanations or invent new classes. This fact prevents us from utilizing traditional metrics relying on confusion matrices. In our approach, we are not guaranteed to receive a miss classification with a false positive label. We cannot apply metrics relying on type I (false positive) and type II (false negative) errors. Therefore, we restrict our evaluation to the calculation of the macro average (unweighted mean). As we receive a free-form text as a response, we apply further preprocessing to extract the predicted label. We count only exact case-insensitive matches. We exclude responses containing additional characters or leading/trailing spaces.

\paragraph*{Implementation}
\label{sec:methods:implementation}
We implemented our approach utilizing Hugging Face \citep{huggingface} for model loading and prediction, and handled data flow and results analysis with Pandas \citep{pandas}. We emphasize that the project is structured to be easily expandable for further LLMs and API integration. We publish our pipeline as a pip repository\footnote{Package available on PyPi: \url{https://pypi.org/project/cltrier_promptClassify}}. The pipeline configuration assumes two main inputs: a list of prompts and a list of models to compare. Each model is queried with each prompt, resulting in multiple experiments. This approach allows for a comprehensive comparison of model performance across different prompts. The querying is performed batch-wise to facilitate efficient and streamlined interactions with the models during the experimentation process. After the querying process, the pipeline uses an automated system for collecting results for each prompt and model combination in every experiment to ensure consistent and reliable data collection. We also include a basic plotting functionality, which assumes a sequential relationship between the two dimensions being analyzed.
%
% -----------------------------
\section{Results}
We utilize local resources to run all experiments. All calculations are performed on a single Nvidia Tesla V100 32GB GPU combined with two Intel Xeon Silver 12 core 2.2GHZ CPUs and 512GB RAM. We developed our experimental environment to run the predictions batch-wise, looping for every model over every prompt.

% -----------------------------
%
\subsection{Methods Comparison}
\begin{figure*}[ht]
    \centering
    \includegraphics[width=\textwidth]{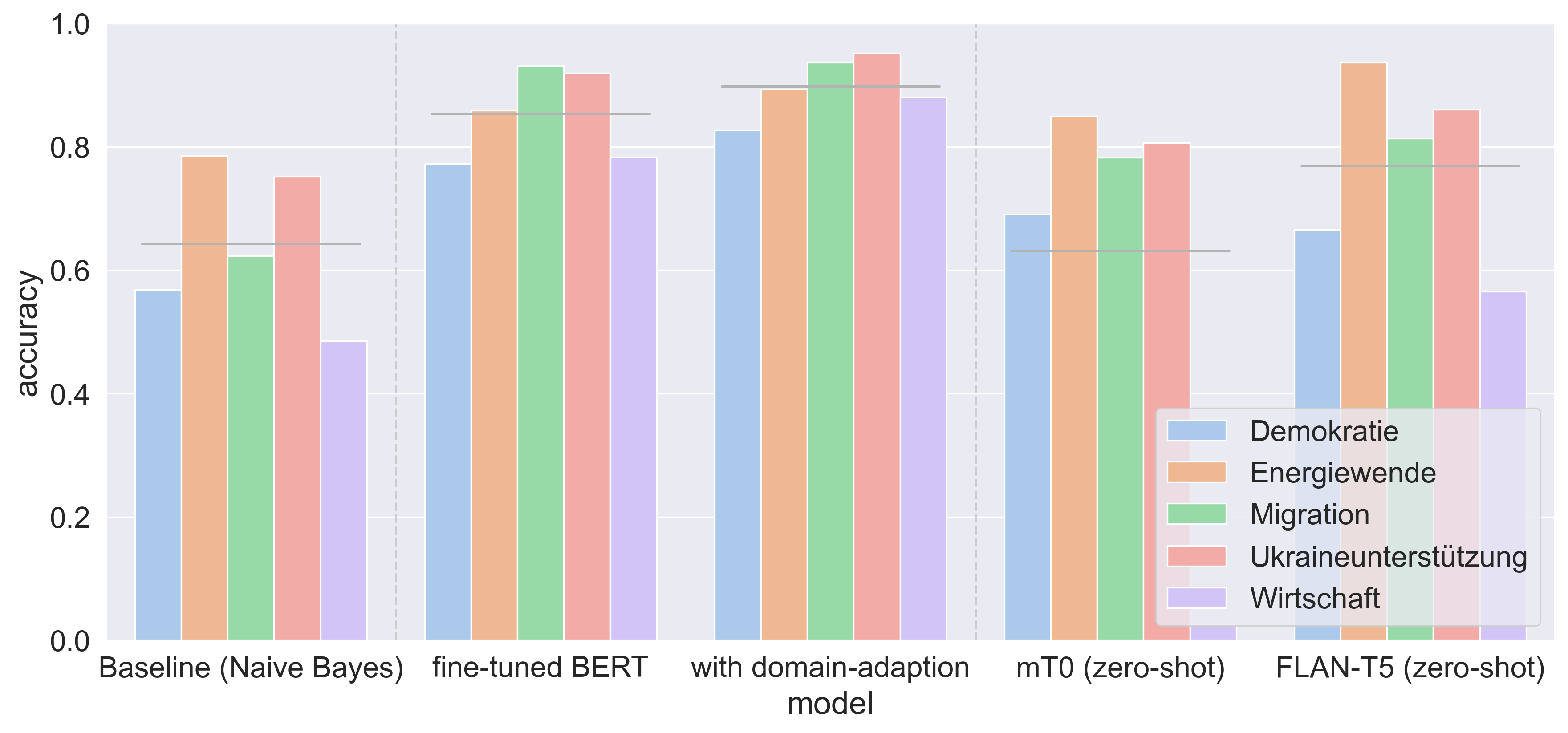}
    \caption{Comparison across methods for topic task}
    \label{graph:results:overview:topic}
\end{figure*}

The comparison between the baseline and fine-tuned transformer models reveals a substantial disparity in their classification performance. While the baseline model achieves an approximate weighted average F1 score of 66\%, the fine-tuned transformer model achieves approximately 86\%, representing a significant difference of 20\% (cmp.~Figure \ref{graph:results:overview:topic}). This contrast emphasizes that the topic possesses an underlying semantic meaning that cannot be effectively captured using a simplistic count-based approach. Instead, the intricate language comprehension capabilities of a transformer model are required to accurately grasp the nuances and subtleties of our topics.

Additionally, we observe variations in performance across different classes for both approaches (cmp.~Table \ref{tab:results:overview:topic}). Both models exhibit lower performance in classifying tweets related to \textquote{Demokratie} (approximately 56\% for baseline vs. 77\% for fine-tuned transformer) and \textquote{Wirtschaft} (approximately 48\% for baseline vs. 78\% for fine-tuned transformer). In contrast, classes with high F1 scores such as \textquote{Energiewende} (approximately 78\% for baseline vs. 85\% for fine-tuned transformer) and \textquote{Ukraineunterstützung} (approximately 75\% for baseline vs. 91\% for fine-tuned transformer) demonstrate superior classification accuracy. We hypothesize that the topics with higher F1 scores possess more distinct and well-defined terminology, making the classification task easier, particularly for the baseline model.

\begin{table*}[ht]
	\centering
	\small
	\begin{tabular}{lr | rr | rr}
		\toprule
		& 
		\makecell[r]{\textbf{Baseline}\\Naive Bayes} & 
		\makecell[r]{\\fine-tuned} & 
		\makecell[r]{\textbf{BERT}\\w/ pre-training} & 
		\makecell[r]{\textbf{mT0}\\zero-shot} & 
		\makecell[r]{\textbf{FLAN-T5}\\zero-shot} 
		\\
		\midrule
		\textbf{Demokratie}            & 0.5684 & 0.7727 & \textbf{0.8276} & 0.6908 & 0.6660          \\
		\textbf{Energiewende}          & 0.7857 & 0.8593 & 0.8939          & 0.8500 & \textbf{0.9368} \\
		\textbf{Migration}             & 0.6230 & 0.9310 & \textbf{0.9367} & 0.7826 & 0.8140          \\
		\textbf{UA-Unterst.}           & 0.7521 & 0.9199 & \textbf{0.9524} & 0.8066 & 0.8604          \\
		\textbf{Wirtschaft}            & 0.4857 & 0.7831 & \textbf{0.8807} & 0.0254 & 0.5657          \\
		\midrule
		\textbf{macro avg}             & 0.6430 & 0.8532 & \textbf{0.8983} & 0.6311 & 0.7686          \\
		\bottomrule
	\end{tabular}
	\caption{Comparison across methods for topic task}
	\label{tab:results:overview:topic}
\end{table*}

% -----------------------------
%
\subsection{Prompting Detail}
Our results show that, with more information, the performance gradually improves with the larger models (cmp.~Table \ref{tab:results:prompting:topic}). However, the smaller versions of each family does not profit from the additional information as they struggle to understand the task description in general, and their responses show that the additional information confuses the model and diffuses the given task. Interestingly, solely mentioning the task name noticeably improves the performance compared to the base prompt. We assume that the information is sufficient for the model to connect inside its internal parametric task memory to a similar task from its own instruction-tuning stage. This provides a glimpse into how zero-shot and in-context learning works within foundation models.

\begin{table*}[ht]
	\centering
	\small
	\begin{tabular}{lrr | rr |rr | rr}
		\toprule
		& \multicolumn{2}{c}{\textbf{base}} 
		& \multicolumn{2}{c}{\textbf{w/ task-name}} 
		& \multicolumn{2}{c}{\textbf{w/ description}} 
		& \multicolumn{2}{c}{\textbf{w/ handbook}} \\
		& \textbf{FLAN-T5} & \textbf{mT0} 
		& \textbf{FLAN-T5} & \textbf{mT0} 
		& \textbf{FLAN-T5} & \textbf{mT0} 
		& \textbf{FLAN-T5} & \textbf{mT0} \\
		\midrule
		\textbf{Demokratie}   & 0.4389 & \textbf{0.7595} & 0.4389          & 0.6832 & 0.5229          & 0.6908 & 0.6660          & 0.0324 \\
		\textbf{Energiewende} & 0.8559 & 0.8015          & 0.8750          & 0.8206 & 0.8588          & 0.8500 & \textbf{0.9368} & 0.6868 \\
		\textbf{Migration}    & 0.9179 & 0.5990          & \textbf{0.9203} & 0.7150 & 0.8865          & 0.7826 & 0.8140          & 0.2126 \\
		\textbf{UA-Unterst.}  & 0.7659 & 0.7000          & 0.8000          & 0.7231 & 0.7330          & 0.8066 & \textbf{0.8604} & 0.6714 \\
		\textbf{Wirtschaft}   & 0.4640 & 0.0000          & 0.4831          & 0.0064 & \textbf{0.6017} & 0.0254 & 0.5657          & 0.1292 \\
		\midrule
		\textbf{macro avg}    & 0.6885 & 0.5720          & 0.7035          & 0.5896 & 0.7206          & 0.6311 & \textbf{0.7686} & 0.3465 \\
		\bottomrule
	\end{tabular}
	\caption{Comparison in prompt detail for topic task}
	\label{tab:results:prompting:topic}
\end{table*}

%
% -----------------------------
\section{Discussion}
While our classification results show comparable performance to the baselines, we observe new challenges unseen in classic machine-learning pipelines. These represent the typical pitfalls of LLMs.

\paragraph*{Hallucinations}
Independently from the sizes both model families fabricate topics not given in our prompts. In particular, the small and base models suffer from this behavior. We place this phenomenon under the term LLM hallucination. In general, it describes the generation of false information when an LLM has no internal information about a task or question asked. Interestingly, the terminology concerning language models and behavior is criticized, and researchers propose the usage of the word confabulation \citep{cha23}. It describes, in a psychiatric context, the behavior of people to invent plausible-sounding justifications that have no basis. These individuals appear to strongly believe in the story and do not intend to deceive with the information \citep{mos95}. The change in terminology and perspective allows for an analysis of the phenomenon in contrast to human behavior and comparison with neural pathologies: \textquote{What are LLMs but humans with extreme amnesia and no central coherence?} \citep{mil23}

\paragraph*{Inconsistencies}
Our results show a highly inconsistent behavior not only between prompt variations but also for different samples and the same prompt. As we described in our results, the models generate responses that do not match our task description, like translation and code snippets for some prompt templates. However, we observe also the occurrence of these phenomena for individual samples while using prompts that provide mostly sound responses. These inconsistencies occur for both models in all sizes, even though mT0 is more affected. Current research investigates negated prompts and shows that models perform significantly worse \citep{jan23}. These results question the task understanding of LLMs and underline how sensitive they are to their inputs. Transferred to our approach, the inconsistencies may be caused by linguistic phenomena inside the Tweets which alters the prompt meaning for the model. 

\paragraph*{Blackbox}
With prompt-based approaches, we overall move more in a direction, where the machine learning black box becomes even more opaque in contrast to traditional ML methods, as we cannot see the prediction scores for each possible class. This is a major disadvantage as optimizing the pipeline relies on comparing the textual results with the provided prompts. Combined with the issue that traditional metrics, which rely on the confusion matrices, are inapplicable, a qualitative analysis during the prompt optimization becomes necessary.
%
% -----------------------------
\section{Conclusion}

Concerning $RQ_1$, our results show that with a well-defined prompt, including a summarized annotation handbook, our prompt-based approach achieves nearly on-par performance with the fine-tuned baseline and surpasses the naive baseline. When taking into account, that we tested a challenging non-English task \textit{in-the-wild,} with restrictions in model and context window size, and the early development stage of freely available instruction-based models, we assume that our results will significantly tilt towards LLMs in the future. Thus, we expect that prompt-based text classification will be highly relevant for future use in academia when empirical studies on large quantities of text are conducted.

Concerning $RQ_2$, analyzing our prompts in detail along the predefined dimension, we found the following: The difference in German and English prompts in the smaller models is especially significant. Only the XL version does understand the German task formulation. Thus, we assume multi-lingual knowledge is reduced significantly during the parameter pruning. Also, we conclude that instruction training on mostly English tasks does not lead to multilingual task generalization despite pre-training the model on multilingual corpora. Despite not understanding the German task description, the models handled German tweets and classes without any issues. That highlights the importance of the prompt formulation and its closeness to tasks seen during the fine-tuning process.

Manipulating the order of the prompt segments shows only a minor impact on the performance. Inserting the full Tweet into the center of the prompt reduces the quality of the results, which highlights the importance of handling long-distance dependencies. Further, the separation between task and content led to confusion due to the usage of symbols possibly resembling programming code.

Concerning the scope of detail, our results show a correlation between the performance and the extent of information provided in the task description. Larger models benefit more from the detailed description. That aligns with current research on the formulation of prompts and model selection for enhancing the quality of prompt-based tasks \citep{whi23}\citep{log21}. In summary, our results support the current techniques for zero-shot prompting proposed in research \citep{liu21} and online learning guides \citep{dai23}.

% -----------------------------
%
\subsection{Future Work}

Our experiments display the SOTA of Mid 2023. The research around LLMs relevant to our approach expands in two dimensions. On a daily base, new models are released larger in size and higher in performance. We highly recommend extending the research to the recent and more capable LMs to harness the full potential of prompt-based annotation. The usage of larger models would not only increase the zero-shot performance but also allow more complex prompt variants \citep{alm23}, \citep{tou23}. We suggest including examples (few-shot) in prompts to improve the results. We expect a reduction of inconsistencies and hallucinations \citep{log21}, coupled with a higher alignment to the annotation intents.

While considering the annotation task \textit{in-the-wild,} it also delivers inconsistencies like human annotations, capturing personal and demographic properties of the annotators might lead to a more insightful annotation outcome. This can be achieved by adding personas to the prompt or conditioning LLMs on individual human behavior. Considering the domain of prompt engineering, the proposal adapts the idea of role prompting, which shapes the output style of the generated text resembling a certain person. This adaptation method significantly enhances the quality and accuracy of generated content \citep{whi23}, \citep{sha23}.

In summary, the potential for mimicking human behavior in text annotation tasks with LLMs seems enormous. While providing computational social science researchers with a powerful new tool, it also opens up many critical uses like personalized opinion manipulation and impersonation. Potentials for abuse have to be closely monitored.

% Bibliography
\bibliography{main}

\end{document}